\documentclass{article}
\PassOptionsToPackage{numbers, sort&compress}{natbib}

\usepackage[preprint]{neurips_2026}

\usepackage[utf8]{inputenc} % allow utf-8 input
\usepackage[T1]{fontenc}    % use 8-bit T1 fonts
\usepackage{graphicx}       % figures
\usepackage{hyperref}       % hyperlinks
\usepackage[nameinlink,capitalize,noabbrev]{cleveref} % smart references
\usepackage{algorithm}      % algorithm float
\usepackage[noend]{algpseudocode}  % pseudocode
\usepackage{url}            % simple URL typesetting
\usepackage{booktabs}       % professional-quality tables
\usepackage{multirow}       % multirow cells in tables
\usepackage{amsfonts}       % blackboard math symbols
\usepackage{nicefrac}       % compact symbols for 1/2, etc.
\usepackage{microtype}      % microtypography
\usepackage{tikz}           % lightweight diagrams
\usetikzlibrary{arrows.meta,positioning}
\usepackage{xcolor}         % colors
\newcommand{\mmpadhighlight}[1]{\textcolor{red!70!black}{\textbf{#1}}}
\crefname{algorithm}{algorithm}{algorithms}
\Crefname{algorithm}{Algorithm}{Algorithms}

\title{Matrix Profile for Time-Series Anomaly Detection:\\A Reproducible Open-Source Benchmark on TSB-AD}

\author{%
  Chin-Chia Michael Yeh \\
  \texttt{myeh003@ucr.edu} \\
}

\begin{document}

\maketitle

\begin{abstract}
Matrix Profile (MP) methods are an interpretable and scalable family of distance-based methods for time-series anomaly detection, but strong benchmark performance still depends on design choices beyond a vanilla nearest-neighbor profile.
This technical report documents an open-source Matrix Profile for Anomaly Detection (MMPAD) submission to TSB-AD, a benchmark that covers both univariate and multivariate time series.
The submitted system combines pre-sorted multidimensional aggregation, efficient exclusion-zone-aware $k$-nearest-neighbor ($k$NN) retrieval for repeated anomalies, and moving-average post-processing.
To serve as a reproducible reference for MP-based anomaly detection on TSB-AD, we detail the released implementation, the hyperparameter settings for the univariate and multivariate tracks, and the corresponding benchmark results.
We further analyze how the system performs on the aggregate leaderboard and across specific dataset characteristics.
The open-source implementation is available at \url{https://github.com/mcyeh/mmpad_tsb}.
\end{abstract}

% Matrix Profile (MP) methods are an interpretable and scalable family of distance-based methods for time-series anomaly detection, but strong benchmark performance still depends on design choices beyond a vanilla nearest-neighbor profile. This technical report documents an open-source Matrix Profile for Anomaly Detection (MMPAD) submission to TSB-AD, a benchmark that covers both univariate and multivariate time series. The submitted system combines pre-sorted multidimensional aggregation, efficient exclusion-zone-aware k-nearest-neighbor (kNN) retrieval for repeated anomalies, and moving-average post-processing. To serve as a reproducible reference for MP-based anomaly detection on TSB-AD, we detail the released implementation, the hyperparameter settings for the univariate and multivariate tracks, and the corresponding benchmark results. We further analyze how the system performs on the aggregate leaderboard and across specific dataset characteristics.The open-source implementation is available at https://github.com/mcyeh/mmpad_tsb.

\section{Introduction}
\label{sec:intro}
Time-series anomaly detection is a fundamental problem because real systems often fail through rare, structured, and temporally localized events~\cite{liu2024elephant}.
Among distance-based approaches, the Matrix Profile (MP) is especially attractive because it is exact, interpretable, and scalable~\cite{yeh2016matrix,yeh2018time,yeh2024matrix}.
It supports anomaly detection through nearest-neighbor discord scoring and has become a widely used reference baseline in benchmark studies~\cite{liu2024elephant}.

This report documents how that basic idea is turned into a reproducible submission on TSB-AD~\cite{liu2024elephant}, a benchmark covering both univariate and multivariate time-series anomaly detection.
The aim is not to propose a new anomaly detector from scratch, but to document the concrete design choices required for a benchmark-ready open-source MP-based system.
In practice, a benchmark-ready submission depends on several ingredients beyond the vanilla one-nearest-neighbor profile, including multidimensional aggregation, efficient $k$-nearest-neighbor ($k$NN) retrieval, and moving-average post-processing.

\begin{figure}[htp]
\centerline{
\includegraphics[width=0.82\linewidth]{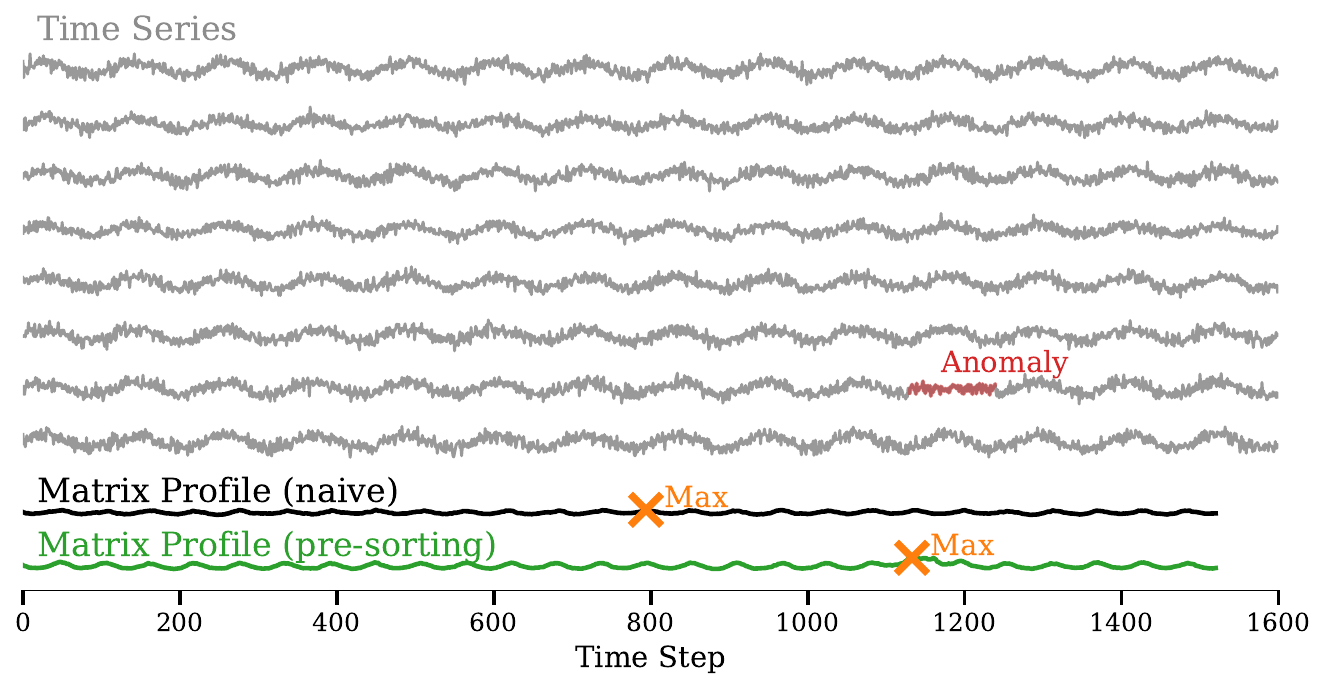}
}
\caption{
Toy illustration of the K-of-N anomaly problem.
The time series contains an anomaly in only one of the eight dimensions.
The naive all-dimension matrix profile reaches its maximum away from the anomalous interval, whereas the pre-sorting matrix profile reaches its maximum near that interval.
The orange X marks the maximum value of each profile.
}
\label{fig:kn_problem}
\end{figure}

A central challenge arises in the multivariate setting.
In the univariate case, comparing all subsequences yields a 2D $n_{\mathrm{sub}} \times n_{\mathrm{sub}}$ pairwise distance matrix.\footnote{For a time series of length $n$ and subsequence length $m$, the number of subsequences is $n_{\mathrm{sub}} = n - m + 1$.}
However, in the $d$-dimensional case, each subsequence pair contributes one distance per dimension, expanding the representation into an $n_{\mathrm{sub}} \times n_{\mathrm{sub}} \times d$ comparison tensor.
The key difficulty is not the pairwise comparison itself, but deciding how to aggregate across that dimension axis to produce an anomaly score.
If an anomalous signal appears in only $K$ out of the $d$ dimensions at a given moment, naive all-dimension aggregation can wash out the anomaly beneath the noise of the remaining $d-K$ normal dimensions.
This creates the K-of-N anomaly problem~\cite{yeh2024matrix}.
\Cref{fig:kn_problem} illustrates this failure mode on a toy eight-dimensional series, where naive aggregation peaks away from the anomalous interval, whereas the pre-sorting profile peaks near it.

As a result, dimensional aggregation is not a minor implementation detail, but one of the central algorithmic decisions in multidimensional MP~\cite{yeh2017matrix,yeh2024matrix}.
This report follows the pre-sorting design, which sorts the dimension-wise comparison scores for each subsequence pair before nearest-neighbor selection~\cite{yeh2017matrix,yeh2024matrix}.
For contrast, the alternative post-sorting design applies dimensional ordering only after dimension-wise nearest neighbors have already been fixed.
This matters because pre-sorting allows dimensional ordering to influence neighbor selection itself, rather than only reordering evidence after the neighbors have already been chosen.
The rationale for adopting pre-sorting, together with a brief contrast against post-sorting, is discussed in \Cref{subsec:aggregation}.
The source paper~\cite{yeh2024matrix} recommends pre-sorting and reports stronger average benchmark performance for the pre-sorting variants it evaluates.

A second key ingredient is efficient $k$NN retrieval.
This matters because recurrent anomalies can defeat one-nearest-neighbor scoring when an anomalous subsequence finds another anomalous subsequence as its closest match.
A major contribution of~\cite{yeh2024matrix} is the efficient exclusion-zone-aware $k$NN algorithm.
Rather than relying on brute-force repeated minimum search or naive full sorting, that algorithm first narrows the candidate set through linear-time selection in the spirit of QuickSelect and introselect~\cite{hoare1961algorithm,musser1997introspective} and then sorts only the reduced candidates~\cite{yeh2024matrix}.
This avoids the extra cost of repeated full-array scans or full sorting and is one reason $k$NN-based matrix-profile scoring remains practical at benchmark scale.

Finally, the submission follows~\cite{yeh2024matrix} in applying moving-average post-processing to the anomaly score curve.
In the implementation studied here, this step is the final score-reduction stage and returns a score vector with the same length as the input time series.
Although simple, it is part of the full pipeline and therefore needs to be documented together with the multidimensional aggregation and $k$NN components.

This report serves both as a method document and as an empirical benchmark study for MP-based anomaly detection on TSB-AD.
Methodologically, it gives a reproducible account of how the released Matrix Profile for Anomaly Detection (MMPAD) implementation turns multidimensional MP into a benchmark-ready detector.
Empirically, it records the benchmark configuration and results of that implementation and then analyzes where MMPAD is competitive on the leaderboard and where it gains or loses ground in the later dataset-characteristic comparisons.
The open-source implementation is available at \url{https://github.com/mcyeh/mmpad_tsb}.
The remainder of the paper is organized as follows.
\Cref{sec:method} details the computational pipeline.
\Cref{sec:experiment} presents the benchmark protocol and results, distilling the main empirical takeaways before \Cref{sec:conclusion} concludes.

\section{Methodology}
\label{sec:method}
This section describes the Matrix Profile for Anomaly Detection (MMPAD) pipeline evaluated on TSB-AD.
For an input time series $X \in \mathbb{R}^{n \times d}$ and subsequence length $m$, let $n_{\mathrm{sub}} = n - m + 1$ denote the number of subsequences.
Throughout this section, we describe the multivariate case, with the univariate case recovered by setting $d = 1$.
We begin with an end-to-end overview and then detail the aggregation, retrieval, and post-processing stages in the same order.
The reported results correspond to the unsupervised evaluation protocol defined by the TSB-AD benchmark~\cite{liu2024elephant}.
Under this protocol, MMPAD is applied to each time series in a self-join fashion.
In this setting, an exclusion zone is used to avoid self-matches and their trivial matches.

\subsection{Pipeline Overview}
\label{subsec:pipeline}
The method can be summarized as a four-stage conceptual pipeline.
These stages summarize how the input time series is transformed into the final anomaly score vector.
Stage 1 computes the dimension-wise pairwise z-normalized Euclidean distances between subsequence pairs.
The resulting pairwise distance tensor has shape $n_{\mathrm{sub}} \times n_{\mathrm{sub}} \times d$, and we denote it by $\mathbf{D}$.
This tensor is a conceptual object used to describe the computation, not a tensor that is explicitly materialized in memory by the implementation.
Stage 2 applies pre-sorting to $\mathbf{D}$.
The purpose and effect of this step are explained in \Cref{subsec:aggregation}.
Stage 3 performs efficient exclusion-zone-aware $k$NN selection on the pre-sorted distances to produce the multidimensional profile tensor $\mathbf{P} \in \mathbb{R}^{n_{\mathrm{sub}} \times d \times k}$, whose axes correspond to subsequence, ordered profile dimension, and neighbor order.
The retrieval step is explained in \Cref{subsec:knn}.
Stage 4 reduces $\mathbf{P}$ to a one-dimensional subsequence score vector $s$ and then applies moving-average smoothing to obtain the final anomaly score vector $y$.
The reduction and smoothing step is explained in \Cref{subsec:postprocessing}.
The overall data flow can be summarized compactly as
\[
X \;\rightarrow\; \mathbf{D} \;\rightarrow\; \mathbf{P} \;\rightarrow\; s \;\rightarrow\; y
\]
The next three subsections explain these aggregation, retrieval, and post-processing stages in detail.

\subsection{Multidimensional Aggregation by Pre-Sorting}
\label{subsec:aggregation}
The first two stages of the pipeline organize pairwise subsequence comparisons into a multidimensional representation that supports anomaly-oriented neighbor selection.
In the univariate case, the matrix profile summarizes a pairwise subsequence comparison matrix by retaining, for each query subsequence, the distance to its nearest match~\cite{yeh2016matrix,yeh2018time}.
In the multivariate case, the pairwise comparison object becomes a tensor because each subsequence pair produces one distance per dimension~\cite{yeh2017matrix,yeh2024matrix}.
If a univariate time series yields an $n_{\mathrm{sub}} \times n_{\mathrm{sub}}$ comparison matrix, then a $d$-dimensional time series yields an $n_{\mathrm{sub}} \times n_{\mathrm{sub}} \times d$ comparison tensor.
In practice, the implementation does not store this full tensor.
Instead, it computes the required pairwise distances for one query subsequence at a time and updates the profile on the fly.

For anomaly detection, naive aggregation across all dimensions can fail.
The reason is the K-of-N anomaly problem introduced in \Cref{sec:intro}: only a subset of dimensions may carry the anomalous signal at a given time, so averaging or summing across all dimensions can hide the anomaly under the remaining normal dimensions~\cite{yeh2024matrix}.
\Cref{fig:kn_problem} in the introduction illustrates this failure mode on a toy example.
The submission therefore follows the pre-sorting strategy introduced for multidimensional MP in~\cite{yeh2017matrix} and adapted to anomaly detection in~\cite{yeh2024matrix}.

\begin{figure}[htp]
\centerline{
\includegraphics[width=0.55\linewidth]{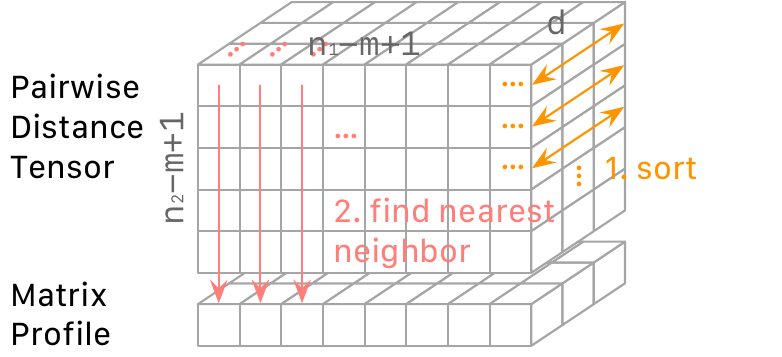}
}
\caption{
The pre-sorting strategy sorts dimension-wise pairwise distances before nearest-neighbor selection in the multidimensional matrix profile.
This ordering lets the sorted dimension-wise distances influence neighbor selection itself, which helps preserve anomalies that appear in only a subset of dimensions.
}
\label{fig:pre_sort}
\end{figure}

\begin{figure}[htp]
\centerline{
\includegraphics[width=0.55\linewidth]{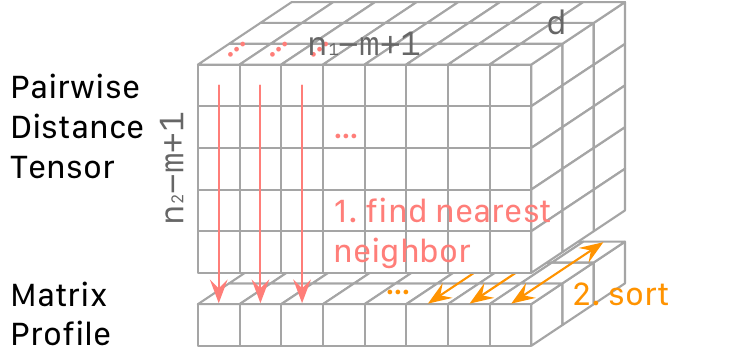}
}
\caption{
The post-sorting strategy sorts dimension-wise matrix-profile values after nearest-neighbor selection.
As a result, dimensional ordering is applied only after the neighbors have already been chosen.
}
\label{fig:post_sort}
\end{figure}

Under pre-sorting, the $d$ dimension-wise distances for each subsequence pair are sorted in the anomaly-oriented order used by the detector.
This local dimensional ranking is performed before nearest-neighbor selection, at the level of subsequence pairs rather than after a multidimensional profile has already been formed.
As described in~\cite{yeh2024matrix}, the $i$-th dimension of the resulting multidimensional profile can be used to detect anomalies that span at least $i$ dimensions.
\Cref{fig:pre_sort} summarizes the pre-sorting design adopted in this report.
The dimension-wise pairwise distances are ordered first, and nearest-neighbor selection is then performed on these ordered distances.
For visual simplicity, \Cref{fig:pre_sort,fig:post_sort} illustrate the one-nearest-neighbor case, so the output is shown as an $n_{\mathrm{sub}} \times d$ matrix rather than a tensor with an additional neighbor axis.

For contrast, \Cref{fig:post_sort} shows the alternative post-sorting design.
There, nearest neighbors are identified independently for each dimension first, and dimensional ordering is applied only afterward.

This ordering difference helps align pre-sorting with the K-of-N anomaly problem.
It allows dimensional ordering to influence neighbor selection itself, rather than being applied only after the nearest neighbors for each dimension have already been chosen.
This is consistent with the source paper's conclusion that the pre-sorting variants it evaluates are generally more desirable for anomaly detection~\cite{yeh2024matrix}.

\subsection{Efficient \texorpdfstring{$k$NN}{kNN} Retrieval}
\label{subsec:knn}
Classical matrix profile scoring uses only the nearest neighbor of each subsequence.
For anomaly detection, this can be brittle when anomalous patterns recur, because an anomalous subsequence may find another anomalous subsequence as its closest match~\cite{yeh2024matrix}.
The submission therefore uses the efficient exclusion-zone-aware $k$NN algorithm proposed in~\cite{yeh2024matrix}.

The key constraint is that $k$NN retrieval in matrix-profile computation must exclude trivial matches~\cite{yeh2016matrix}.
Once a neighbor is selected, nearby subsequences within the exclusion zone must be ignored when searching for later neighbors.
As a result, the $k$-th valid neighbor is not simply the $k$-th element of the raw distance array, so a single standard linear-time selection routine such as QuickSelect or introselect cannot be applied directly~\cite{hoare1961algorithm,musser1997introspective,yeh2024matrix}.
The naive alternatives are repeated minimum search with exclusion after each accepted neighbor or full sorting followed by exclusion checks, both of which become inefficient at benchmark scale~\cite{yeh2024matrix}.
To describe the selection kernel in a join-agnostic way, let $n_{\mathrm{query}}$ and $n_{\mathrm{ref}}$ denote the numbers of query and reference subsequences, respectively.
In the self-join case, both reduce to $n_{\mathrm{sub}}$.

The algorithm used here addresses that issue by first applying linear-time selection to keep only a reduced candidate set before sorting~\cite{hoare1961algorithm,musser1997introspective}.
Following the presentation in~\cite{yeh2024matrix}, this reduced set has size $k m$.
In the implementation, the same idea is parameterized by the explicit exclusion-zone length $\ell_{\mathrm{ex}} = \lfloor 0.5 m \rfloor$, so the reduced candidate count remains proportional to $k m$ under the default self-join setting.
It then sorts only that reduced candidate set rather than the full set of $n_{\mathrm{ref}}$ candidates~\cite{yeh2024matrix}.
When $n_{\mathrm{ref}}$ is much larger than $k m \log(k m)$, this yields the same overall scaling reported in~\cite{yeh2024matrix}: $O(n_{\mathrm{query}} n_{\mathrm{ref}} d)$ instead of $O(k n_{\mathrm{query}} n_{\mathrm{ref}} d)$ for brute force or $O(n_{\mathrm{query}} n_{\mathrm{ref}} \log n_{\mathrm{ref}} d)$ for naive sorting.
This algorithmic improvement is a key reason the $k$NN extension in~\cite{yeh2024matrix} reduces the asymptotic cost of multidimensional matrix-profile scoring.
\Cref{alg:knn_fast} summarizes the inner $k$-th-neighbor selection kernel used in the submission.

\begin{algorithm}[htp]
\centering
\caption{Efficient exclusion-zone-aware $k$NN selection}
\label{alg:knn_fast}
\small
\begin{algorithmic}[1]
\Require Pairwise score tensor $\mathbf{D} \in \mathbb{R}^{n_{\mathrm{query}} \times n_{\mathrm{ref}} \times d}$, number of neighbors $k \in \mathbb{N}$, exclusion-zone length $\ell_{\mathrm{ex}} \in \mathbb{N}$
\Ensure $k$-th neighbor index matrix $I \in \mathbb{N}^{n_{\mathrm{query}} \times d}$
\Function{Find$k$NN}{$\mathbf{D}, k, \ell_{\mathrm{ex}}$}
\State $I \gets$ zero matrix with size $n_{\mathrm{query}} \times d$
\For{$i \gets 1$ to $n_{\mathrm{query}}$}
    \For{$j \gets 1$ to $d$}
        \State $\mathbf{i}_{\mathrm{neighbors}} \gets \textsc{ArgSelect}(\mathbf{D}[i, :, j], \min(2 k \ell_{\mathrm{ex}},\; n_{\mathrm{ref}} - 1))$ \Comment{$O(n_{\mathrm{ref}})$}
        \State $\mathbf{i}_{\mathrm{order}} \gets \textsc{ArgSort}(\mathbf{D}[i, \mathbf{i}_{\mathrm{neighbors}}, j])$ \Comment{$O(k \ell_{\mathrm{ex}} \log(k \ell_{\mathrm{ex}}))$}
        \State $\mathbf{i}_{\mathrm{neighbors}} \gets \mathbf{i}_{\mathrm{neighbors}}[\mathbf{i}_{\mathrm{order}}]$
        \State $l \gets 0$
        \ForAll{$i_{\mathrm{neighbor}} \in \mathbf{i}_{\mathrm{neighbors}}$} \Comment{$O(k \ell_{\mathrm{ex}})$}
            \If{$i_{\mathrm{neighbor}}$ is a trivial match of a previously accepted neighbor}
                \State \textbf{continue}
            \EndIf
            \State $l \gets l + 1$
            \If{$l = k$}
                \State \textbf{break}
            \EndIf
        \EndFor
        \State $I[i, j] \gets i_{\mathrm{neighbor}}$
    \EndFor
\EndFor
\State \Return $I$ \Comment{Overall: $O(n_{\mathrm{query}} n_{\mathrm{ref}} d)$ when $n_{\mathrm{ref}}$ dominates $k \ell_{\mathrm{ex}} \log(k \ell_{\mathrm{ex}})$}
\EndFunction
\end{algorithmic}
\end{algorithm}

For each subsequence and each ordered profile dimension, the retrieval step records the first through $k$-th neighbors, yielding the multidimensional profile tensor $\mathbf{P}$.
This tensor is the input to the post-processing stage.

\subsection{Post-Processing}
\label{subsec:postprocessing}
The output of the retrieval stage is the profile tensor $\mathbf{P}$, not yet the final anomaly score curve used for evaluation.
This tensor must be reduced to a one-dimensional score sequence before benchmark evaluation.
Let $d^\ast$ denote the selected cutoff on the ordered profile-dimension axis and $k^\ast$ denote the selected neighbor order.
This reduction can be viewed in three stages.

First, it keeps only the first $d^\ast$ dimensions and the first $k^\ast$ neighbors.
Second, if a later dimension or later neighbor contains a non-finite value, the implementation backs off to the preceding valid dimension or preceding valid neighbor.
This prevents exclusion-induced missing values from propagating into the final score.
Third, it takes the score at ordered profile dimension $d^\ast$ and neighbor order $k^\ast$ as the subsequence anomaly score and flips the sign so that larger values correspond to more anomalous behavior.

At this point, the score vector still has length $n_{\mathrm{sub}}$ because it is defined over subsequences rather than timestamps.
Following~\cite{yeh2024matrix}, the submission then applies moving-average post-processing to the anomaly score curve.
In the implementation, the subsequence score vector is padded with $m-1$ NaN values on both ends, and a sliding mean over windows of length $m$ is then computed while ignoring the padded NaN values.
This produces a final anomaly score vector of length $n$, which matches the length of the input time series.
Operationally, each timestamp-level score is the mean of the subsequence scores whose windows cover that timestamp, which smooths the final anomaly curve.
\Cref{alg:score_reduction} summarizes the post-processing pipeline used after multidimensional $k$NN retrieval.

\begin{algorithm}[htp]
\centering
\caption{Post-processing}
\label{alg:score_reduction}
\small
\begin{algorithmic}[1]
\Require Profile tensor $\mathbf{P} \in \mathbb{R}^{n_{\mathrm{sub}} \times n_{\mathrm{dim}} \times n_{\mathrm{neighbor}}}$, selected dimension $d^\ast$, selected neighbor order $k^\ast$, subsequence length $m$
\Ensure Final anomaly score vector $y \in \mathbb{R}^{n}$
\Function{PostProcess}{$\mathbf{P}, d^\ast, k^\ast, m$}
\State $\mathbf{P} \gets \mathbf{P}[:, 1:d^\ast, 1:k^\ast]$
\For{$i \gets 2$ to $d^\ast$}
    \State Replace non-finite entries in $\mathbf{P}[:, i, :]$ with the corresponding entries in $\mathbf{P}[:, i - 1, :]$
\EndFor
\State $\mathbf{S} \gets \mathbf{P}[:, d^\ast, :]$
\For{$j \gets 2$ to $k^\ast$}
    \State Replace non-finite entries in $\mathbf{S}[:, j]$ with the corresponding entries in $\mathbf{S}[:, j - 1]$
\EndFor
\State $s \gets -\mathbf{S}[:, k^\ast]$
\State $\tilde{s} \gets$ \textsc{PadNaN}$(s, m - 1, m - 1)$
\For{$t \gets 1$ to $n$}
    \State $y_t \gets$ \textsc{NaNMean}$(\tilde{s}[t:t + m - 1])$
\EndFor
\State \Return $y$
\EndFunction
\end{algorithmic}
\end{algorithm}

In the implementation, the selected cutoff $d^\ast$ can be specified either as an integer or as a fraction of the total dimensionality, in which case the fraction is converted to an integer by ceiling.
This allows the benchmark configuration to search over either absolute dimensional cutoffs or proportional dimensional cutoffs without changing the core method.

The implementation also supports optional budget-aware downsampling for long sequences.
When that path is used during fitting, the working series is repeatedly downsampled by a factor of two by keeping every other timestamp until a proxy runtime cost falls below the configured time budget.
After each reduction, the subsequence length is re-inferred on the downsampled series.
The detector is then run on this downsampled working series, and the final anomaly score vector is linearly interpolated back to the original sequence length before evaluation.
This preserves the benchmark evaluation interface while allowing the implementation to respect a fixed runtime budget.
Under the released default time budget used in the reported experiments, this path is triggered for no univariate evaluation files and for one multivariate evaluation file.
With the full MMPAD pipeline now defined, we turn to its empirical evaluation on the TSB-AD benchmark.

\section{Experiment}
\label{sec:experiment}
This section documents the benchmark protocol, the hyperparameter selection procedure, the evaluation-set results, and the performance of MMPAD across dataset characteristics on TSB-AD.

\subsection{Evaluation Protocol}
The results reported here follow the unsupervised evaluation protocol defined by TSB-AD~\cite{liu2024elephant,liu2026tsbadweb}.
The official evaluation lists contain 350 univariate time series and 180 multivariate time series, while the official tuning lists contain 48 univariate time series and 20 multivariate time series~\cite{liu2024elephant,liu2026tsbadweb}.
Within that protocol, the benchmark reports separate univariate and multivariate tracks.
We use the tuning split only for hyperparameter selection and reserve the evaluation split for the final benchmark comparison.

The benchmark documentation states that the evaluation results use time series that are z-score normalized by default, and the official runners evaluate each score sequence with the released metric implementation~\cite{liu2026tsbadweb,liu2026tsbadrepo}.
In those runners, the sliding-window length used by the range-aware metrics is inferred from the first dimension by the same autocorrelation-based period estimator used elsewhere in the benchmark codebase~\cite{liu2026tsbadrepo}.

TSB-AD reports nine metrics in total: AUC-PR, AUC-ROC, VUS-PR, VUS-ROC, Standard-F1, PA-F1, Event-based-F1, R-based-F1, and Affiliation-F.
This report focuses on VUS-PR because the benchmark paper identifies it as the primary comparison metric~\cite{liu2024elephant}.

We summarize performance in two ways.
The first is the average VUS-PR over the evaluation files.
The second is an average per-dataset VUS-PR rank.
To compute that second summary, we take the released per-dataset VUS-PR tables~\cite{liu2026tsbadrepo}, add the MMPAD settings studied here, rank methods by VUS-PR within each dataset, and then average those ranks across datasets, so lower is better.

\subsection{Hyperparameter Tuning}
The submission includes two selected hyperparameter settings.
One is chosen by maximizing average VUS-PR on the tuning split, and the other is chosen by minimizing the average per-dataset VUS-PR rank on the same split.
The subsequence length $m$ is not tuned as a free hyperparameter in these runs.
Instead, it is inferred from the first dimension by the same autocorrelation-based period estimator used to set the evaluation window in the benchmark runners.
For the univariate track, the main hyperparameter search varies $k \in \{1,5,10,15\}$.
For the multivariate track, it additionally varies $d^\ast \in \{1,0.1,0.3,0.5,0.7\}$.
When $d^\ast < 1$, it is interpreted as a fraction of the total dimensionality and converted to an integer cutoff by ceiling.
The two tracks differ because dimensional selection is meaningful only in the multivariate case.

\Cref{tab:hp_selection} shows the resulting selections together with their average tuning-split VUS-PR and average tuning-split rank.
In the univariate track, the two criteria differ only in the selected neighbor count.
The VUS-PR criterion selects $k=5$, while the rank criterion selects $k=10$.
In the multivariate track, both criteria agree on $k=15$ and differ only in the selected dimensional cutoff.
The VUS-PR criterion selects $d^\ast=0.7$, while the rank criterion selects $d^\ast=0.5$.
This agreement shows that the stronger submission settings are not based on the vanilla one-nearest-neighbor profile.
Throughout these experiments, pre-sorting and moving-average post-processing remain fixed design choices, so the tuning evidence reported below pertains only to $k$ and, in the multivariate track, $d^\ast$.
Because VUS-PR is the benchmark's primary comparison metric~\cite{liu2024elephant}, the remaining experiment subsections focus on the submission setting selected by average VUS-PR on the tuning split.

\begin{table}[htp]
\centering
\small
\caption{Hyperparameter selections from the tuning split. The VUS-PR and Rank columns report the average tuning-split VUS-PR and the average tuning-split per-dataset VUS-PR rank.}
\label{tab:hp_selection}
\begin{tabular}{@{}llp{0.32\linewidth}rr@{}}
\toprule
Track & Rule & HP & VUS-PR & Rank \\
\midrule
Univariate & VUS-selected & k=5 & 0.4256 & 4.9688 \\
Univariate & rank-selected & k=10 & 0.4038 & 4.8958 \\
Multivariate & VUS-selected & d=0.7, k=15 & 0.3879 & 23.1750 \\
Multivariate & rank-selected & d=0.5, k=15 & 0.3597 & 19.7000 \\
\bottomrule
\end{tabular}
\end{table}

\subsection{Main Benchmark Results}
\Cref{tab:leaderboard} focuses on the submission configuration selected on the tuning split by average VUS-PR.
For the benchmark baselines, both the average VUS-PR values and the rank column are recomputed from the released per-dataset VUS-PR tables~\cite{liu2026tsbadrepo} after adding the MMPAD setting studied here.
In the univariate track, MMPAD reaches an average VUS-PR of 0.4399, ahead of Sub-PCA at 0.4234.
Its average per-dataset rank is 12.5943, which is second only to POLY at 12.4971.
This places MMPAD at the top of the univariate benchmark comparison by mean VUS-PR, even though POLY remains slightly stronger by average per-dataset rank.

In the multivariate track, MMPAD achieves the highest average VUS-PR at 0.3539, ahead of CNN at 0.3130.
Its average per-dataset rank is 10.7889, indicating that these gains are concentrated on part of the benchmark rather than being uniform across all datasets.
This contrast between average score and average rank motivates the dataset-level comparison below.

\begin{table}[htp]
\centering
\small
\caption{Top-10 benchmark comparison in each track, showing MMPAD together with the nine strongest baselines by average VUS-PR and average per-dataset VUS-PR rank.}
\label{tab:leaderboard}
\begin{tabular}{@{}clrr@{}}
\toprule
\multicolumn{1}{@{}c}{} & Method & VUS-PR ($\uparrow$) & Rank ($\downarrow$) \\
\midrule
\multirow{10}{*}{\rotatebox[origin=c]{270}{Univariate}} & \mmpadhighlight{MMPAD} & \mmpadhighlight{0.4399} & \mmpadhighlight{12.5943} \\
 & Sub-PCA & 0.4234 & 13.7343 \\
 & KShapeAD & 0.4008 & 14.9029 \\
 & POLY & 0.3898 & 12.4971 \\
 & Series2Graph & 0.3881 & 13.5429 \\
 & MOMENT (FT) & 0.3857 & 13.8300 \\
 & MOMENT (ZS) & 0.3790 & 14.5229 \\
 & KMeansAD & 0.3664 & 14.9614 \\
 & USAD & 0.3612 & 16.2014 \\
 & Sub-KNN & 0.3501 & 14.9186 \\
\midrule
\multirow{10}{*}{\rotatebox[origin=c]{270}{Multivariate}} & \mmpadhighlight{MMPAD} & \mmpadhighlight{0.3539} & \mmpadhighlight{10.7889} \\
 & CNN & 0.3130 & 7.6778 \\
 & OmniAnomaly & 0.3122 & 10.3917 \\
 & PCA & 0.3096 & 9.7972 \\
 & LSTMAD & 0.3066 & 8.7000 \\
 & USAD & 0.3041 & 10.6583 \\
 & AutoEncoder & 0.2950 & 10.9472 \\
 & KMeansAD & 0.2949 & 8.6611 \\
 & CBLOF & 0.2731 & 11.4611 \\
 & MCD & 0.2711 & 11.7944 \\
\bottomrule
\end{tabular}
\end{table}

\subsection{Performance by Dataset Characteristics}
To look beyond the aggregate leaderboard, \Cref{tab:characteristics} regroups the evaluation data by coarse dataset characteristics: the number of anomaly segments, point versus sequence anomalies, series length, anomaly ratio, and anomaly duration.
These tables are meant to answer a narrower question than the leaderboard: against the strongest fixed baseline in each track, on what kinds of datasets does MMPAD gain or lose ground?

The membership of each setting is determined directly from the dataset-level annotations and summary statistics included in the released per-dataset TSB-AD benchmark tables~\cite{liu2026tsbadrepo}.
Single versus multiple anomaly is defined by whether the benchmark metadata records one anomaly segment or more than one.
Point versus sequence anomaly follows the benchmark's anomaly-type label for each dataset.
Shorter versus longer series, low versus high anomaly ratio, and short versus long anomaly duration are each defined by a track-wise median threshold.
These settings are overlapping attribute-based groups rather than one mutually exclusive partition, so a dataset can, for example, be simultaneously single-anomaly, sequence-anomaly, low-ratio, and long-series.
For the univariate track, the median thresholds are series length $= 18227.5$, anomaly ratio $= 0.0215$, and average anomaly duration $= 116.2292$.
For the multivariate track, the corresponding thresholds are series length $= 46655$, anomaly ratio $= 0.0375$, and average anomaly duration $= 241.25$.
For the separate multivariate dimensionality breakdown discussed later in this subsection, the datasets are grouped into quartile-like bins by channel count, which yields the four observed ranges 2--3, 4--19, 20--31, and 32--248 dimensions.

In each track, we compare MMPAD against the strongest non-MMP baseline in the aggregate benchmark table above.
That baseline is Sub-PCA in the univariate track and CNN in the multivariate track.
Positive mean gaps indicate that MMPAD is, on average, better than that fixed rival within the subset.
This fixed-rival design is meant to provide a stable subgroup comparison, not to decompose the average-rank metric from \Cref{tab:leaderboard}.

\begin{table}[htp]
\centering
\small
\caption{Performance by dataset characteristic. The univariate comparison is MMPAD versus Sub-PCA, and the multivariate comparison is MMPAD versus CNN. Positive gaps favor MMPAD. Short/long series, short/long anomalies, and low/high anomaly ratio are defined by track-wise median thresholds.}
\label{tab:characteristics}
\begin{tabular}{@{}lrrrrrr@{}}
\toprule
& \multicolumn{3}{c}{Univariate} & \multicolumn{3}{c}{Multivariate} \\
\cmidrule(lr){2-4}\cmidrule(lr){5-7}
Split & Gap & Win rate & \# series & Gap & Win rate & \# series \\
\midrule
Single anomaly & +0.1603 & 0.6398 & 161 & -0.0809 & 0.3617 & 47 \\
Multiple anomalies & -0.1060 & 0.4233 & 189 & +0.0840 & 0.4962 & 133 \\
Point anomaly & +0.2382 & 0.7143 & 49 & -0.2006 & 0.0000 & 13 \\
Sequence anomaly & -0.0231 & 0.4876 & 322 & +0.0409 & 0.4611 & 180 \\
Short series & +0.1112 & 0.5943 & 175 & -0.1331 & 0.2308 & 91 \\
Long series & -0.0782 & 0.4514 & 175 & +0.2189 & 0.6966 & 89 \\
Low anomaly ratio & +0.2225 & 0.7314 & 175 & +0.0658 & 0.5111 & 90 \\
High anomaly ratio & -0.1896 & 0.3143 & 175 & +0.0160 & 0.4111 & 90 \\
Short anomalies & +0.1857 & 0.6629 & 175 & -0.0438 & 0.2778 & 90 \\
Long anomalies & -0.1528 & 0.3829 & 175 & +0.1256 & 0.6444 & 90 \\
\bottomrule
\end{tabular}
\end{table}

The univariate and multivariate tracks show different patterns.
In the univariate track, relative to Sub-PCA, MMPAD is strongest on single-anomaly, point-anomaly, low-anomaly-ratio, short-anomaly, and shorter-series subsets, and it loses ground on multiple-anomaly, sequence-anomaly, high-anomaly-ratio, long-anomaly, and longer-series subsets.
These results suggest that broader and more repeated anomalous behavior remains a useful univariate target for future variants.
These univariate patterns highlight an open opportunity to test whether compact dictionary representations of time series~\cite{yeh2022error} can be incorporated into the current discord pipeline, allowing the system to better capture broader, more recurrent anomalous behaviors, such as those found in the multiple-anomaly and long-anomaly subsets.

In the multivariate track, relative to CNN, MMPAD is strongest on longer-series, longer-anomaly, and multiple-anomaly subsets, and it is weakest on single-anomaly, point-anomaly, shorter-series, and shorter-anomaly subsets.
The aggregate leaderboard adds useful context.
Among the strongest multivariate baselines in \Cref{tab:leaderboard} are CNN, LSTMAD, OmniAnomaly, PCA, USAD, and AutoEncoder, all of which rely on forecasting, reconstruction, or subspace modeling rather than only local subsequence matching.
Taken together, these results suggest that factors beyond local subsequence matching may contribute to the remaining multivariate gaps.
In other words, pre-sorting addresses the K-of-N aggregation problem, but it does not by itself remove all of the multivariate weaknesses observed on TSB-AD-M.
The same contrast appears when the evaluation datasets are grouped by dimensionality.
Relative to CNN, the mean MMPAD gap is:
\begin{itemize}
\item 2--3 dimensions: $+0.2517$.
\item 4--19 dimensions: $+0.0798$.
\item 20--31 dimensions: $-0.0801$.
\item 32--248 dimensions: $-0.1724$.
\end{itemize}
This trend suggests that the multivariate advantage of MMPAD is concentrated on lower-dimensional datasets and erodes as dimensionality increases.
A promising direction for future work is applying PCA whitening~\cite{wikipedia2025whitening} prior to the MMPAD pipeline, providing a direct way to test whether decorrelating and rescaling cross-channel structures successfully narrows the remaining gaps in the point-anomaly and short-anomaly subsets.

\subsection{CPU--GPU Comparison}
This subsection is included only as an implementation-consistency check between the open-source CPU and GPU versions.
At the aggregate level, the two implementations produce very similar benchmark results.
In the univariate track, the average VUS-PR difference is below 0.0001, and the largest absolute difference on any evaluation dataset is 0.0055.
The same conclusion holds in the multivariate track.
The average GPU$-$CPU difference is only 0.0008, with a mean absolute per-dataset difference of 0.0024 and a maximum absolute difference of 0.0692.
Taken together, these results provide a useful implementation check: the CPU and GPU versions agree closely in aggregate, even though some individual datasets differ more noticeably.

\subsection{Key Takeaways}
\begin{itemize}
\item Matrix-profile-based anomaly detection remains highly competitive on TSB-AD.
MMPAD is first on both the univariate and multivariate leaderboards by average VUS-PR.
\item The tuning results consistently prefer $k>1$.
Benchmark-ready MMPAD is therefore more than the vanilla one-nearest-neighbor profile.
\item The univariate results by dataset characteristics suggest that broader, longer, and repeated anomalies remain a useful target for future variants.
A promising direction for future research is testing whether compact dictionary representations of time series~\cite{yeh2022error} can augment the current discord pipeline, allowing for a direct re-evaluation of the high-ratio and long-anomaly subsets.
\item The multivariate results suggest a different direction.
The multivariate leaderboard and the dimensionality breakdown both point to cross-channel structure as a plausible contributor to the remaining multivariate gaps.
To address this, an open avenue for future work is applying PCA whitening~\cite{wikipedia2025whitening} prior to MMPAD, providing a direct way to test whether decorrelating cross-channel structures improves performance on the point-anomaly and short-anomaly subsets.
\end{itemize}

\section{Conclusion}
\label{sec:conclusion}
This report documented the MMPAD submission to TSB-AD from method design through benchmark evaluation.
MMPAD is first on both the univariate and multivariate leaderboards by average VUS-PR.
The dataset-characteristic analysis also highlights two open challenges for future MP-based anomaly detection.
In the univariate track, one recurring gap relative to a strong subspace baseline is broader or repeated anomalous behavior, which suggests testing whether compact dictionary representations of time series~\cite{yeh2022error} can augment the current discord pipeline.
In the multivariate track, weaker results on point-anomaly, short-anomaly, and higher-dimensional subsets suggest testing preprocessing that makes normal cross-channel structure more explicit, such as PCA whitening before running MMPAD~\cite{wikipedia2025whitening}.

\bibliographystyle{abbrvnat}
\bibliography{section/reference}

@inproceedings{yeh2024matrix,
  title={Matrix profile for anomaly detection on multidimensional time series},
  author={Yeh, Chin-Chia Michael and Der, Audrey and Saini, Uday Singh and Lai, Vivian and Zheng, Yan and Wang, Junpeng and Dai, Xin and Zhuang, Zhongfang and Fan, Yujie and Chen, Huiyuan and others},
  booktitle={2024 IEEE International Conference on Data Mining (ICDM)},
  pages={911--916},
  year={2024},
  organization={IEEE}
}

@inproceedings{yeh2017matrix,
  title={Matrix profile VI: Meaningful multidimensional motif discovery},
  author={Yeh, Chin-Chia Michael and Kavantzas, Nickolas and Keogh, Eamonn},
  booktitle={2017 IEEE international conference on data mining (ICDM)},
  pages={565--574},
  year={2017},
  organization={IEEE}
}

@article{liu2024elephant,
  title={The elephant in the room: Towards a reliable time-series anomaly detection benchmark},
  author={Liu, Qinghua and Paparrizos, John},
  journal={Advances in Neural Information Processing Systems},
  volume={37},
  pages={108231--108261},
  year={2024}
}

@misc{liu2026tsbadweb,
  title={The Elephant in the Room: Towards A Reliable Time-series Anomaly Detection Benchmark},
  author={Liu, Qinghua and Paparrizos, John},
  year={2026},
  howpublished={\url{https://thedatumorg.github.io/TSB-AD/}},
  note={Benchmark website, last updated April 1, 2026, accessed April 1, 2026}
}

@misc{liu2026tsbadrepo,
  title={TSB-AD benchmark repository},
  author={{TheDatumOrg}},
  year={2026},
  howpublished={\url{https://github.com/TheDatumOrg/TSB-AD}},
  note={GitHub repository, accessed April 2, 2026}
}

@inproceedings{yeh2016matrix,
  title={Matrix profile I: all pairs similarity joins for time series: a unifying view that includes motifs, discords and shapelets},
  author={Yeh, Chin-Chia Michael and Zhu, Yan and Ulanova, Liudmila and Begum, Nurjahan and Ding, Yifei and Dau, Hoang Anh and Silva, Diego Furtado and Mueen, Abdullah and Keogh, Eamonn},
  booktitle={2016 IEEE 16th international conference on data mining (ICDM)},
  pages={1317--1322},
  year={2016},
  organization={Ieee}
}

@article{musser1997introspective,
  title={Introspective sorting and selection algorithms},
  author={Musser, David R},
  journal={Software: Practice and Experience},
  volume={27},
  number={8},
  pages={983--993},
  year={1997},
  publisher={Wiley Online Library}
}

@article{hoare1961algorithm,
  title={Algorithm 65: find},
  author={Hoare, Charles AR},
  journal={Communications of the ACM},
  volume={4},
  number={7},
  pages={321--322},
  year={1961},
  publisher={ACM New York, NY, USA}
}

@article{yeh2018time,
  title={Time series joins, motifs, discords and shapelets: a unifying view that exploits the matrix profile},
  author={Yeh, Chin-Chia Michael and Zhu, Yan and Ulanova, Liudmila and Begum, Nurjahan and Ding, Yifei and Dau, Hoang Anh and Zimmerman, Zachary and Silva, Diego Furtado and Mueen, Abdullah and Keogh, Eamonn},
  journal={Data Mining and Knowledge Discovery},
  volume={32},
  number={1},
  pages={83--123},
  year={2018},
  publisher={Springer}
}

@misc{wikipedia2025whitening,
  title={Whitening transformation},
  author={{Wikipedia contributors}},
  year={2025},
  howpublished={\url{https://en.wikipedia.org/wiki/Whitening_transformation}},
  note={Wikipedia, last edited August 28, 2025, accessed April 2, 2026}
}

@inproceedings{yeh2022error,
  title={Error-bounded approximate time series joins using compact dictionary representations of time series},
  author={Yeh, Chin-Chia Michael and Zheng, Yan and Wang, Junpeng and Chen, Huiyuan and Zhuang, Zhongfang and Zhang, Wei and Keogh, Eamonn},
  booktitle={Proceedings of the 2022 SIAM International Conference on Data Mining (SDM)},
  pages={181--189},
  year={2022},
  organization={SIAM}
}

\end{document}